\documentclass[letterpaper, 10 pt, conference]{ieeeconf}  

\IEEEoverridecommandlockouts                              

\usepackage{graphics} 
\usepackage{epsfig} 
\usepackage{mathptmx} 
\usepackage{times} 
\usepackage{amsmath} 
\usepackage{amssymb}  
\usepackage{algorithm}
\usepackage{algpseudocode}
\usepackage{gensymb}
\usepackage{multirow}
\usepackage{subcaption}
\usepackage{cite}

\makeatletter
\let\NAT@parse\undefined
\makeatother
\usepackage{hyperref}

\usepackage{color}

\title{\LARGE \bf
Deep Online Correction for Monocular Visual Odometry
}

\author{Jiaxin Zhang$^{1}$, Wei Sui$^{1}$, Xinggang Wang$^{2}$, Wenming
Meng$^{1}$, Hongmei Zhu$^{1}$, Qian Zhang$^{1}$\\$^{1}$Horizon Robotics
$^{2}$School of EIC, Huazhong University of Science and Technology\\{\tt\small
\{jiaxin02.zhang, wei.sui, wenming.meng, hongmei.zhu,,
qian01.zhang\}@horizon.ai},\\\tt\small{xgwang@hust.edu.cn}
}

\begin{document}

\maketitle
\thispagestyle{empty}
\pagestyle{empty}

\begin{abstract}
	In this work, we propose a novel deep online correction (DOC) framework for
	monocular visual odometry. The whole pipeline has two stages: First, depth
	maps and initial poses are obtained from convolutional neural networks
	(CNNs) trained in self-supervised manners. Second, the poses predicted by
	CNNs are further improved by minimizing photometric errors via gradient
	updates of poses during inference phases. The benefits of our proposed
	method are twofold: 1) Different from online-learning methods, DOC does not
	need to calculate gradient propagation for parameters of CNNs. Thus, it
	saves more computation resources during inference phases. 2) Unlike hybrid
	methods that combine CNNs with traditional methods, DOC fully relies on deep
	learning (DL) frameworks. Though without complex back-end optimization
	modules, our method achieves outstanding performance with relative transform
	error (RTE) = 2.0\% on KITTI Odometry benchmark for Seq. 09, which
	outperforms traditional monocular VO frameworks and is comparable to hybrid
	methods.
\end{abstract}

\section{INTRODUCTION}

Monocular visual odometry has attracted more and more attention for its wide
applications in robotics, autonomous driving and augmented reality (AR). As an
effective complement to other sensors such as GPS, Inertial Navigation System
(INS) and wheel odometry, etc., monocular VO is popular for its low cost and
easy access.

A larger number of robust and accurate monocular VO systems have been developed
in the past decades \cite{tutorialVO}. These methods can be roughly classified
into three categories: traditional, DL-based and hybrid methods.
\begin{figure}[thpb]
	\centerline{\includegraphics[width=90mm]{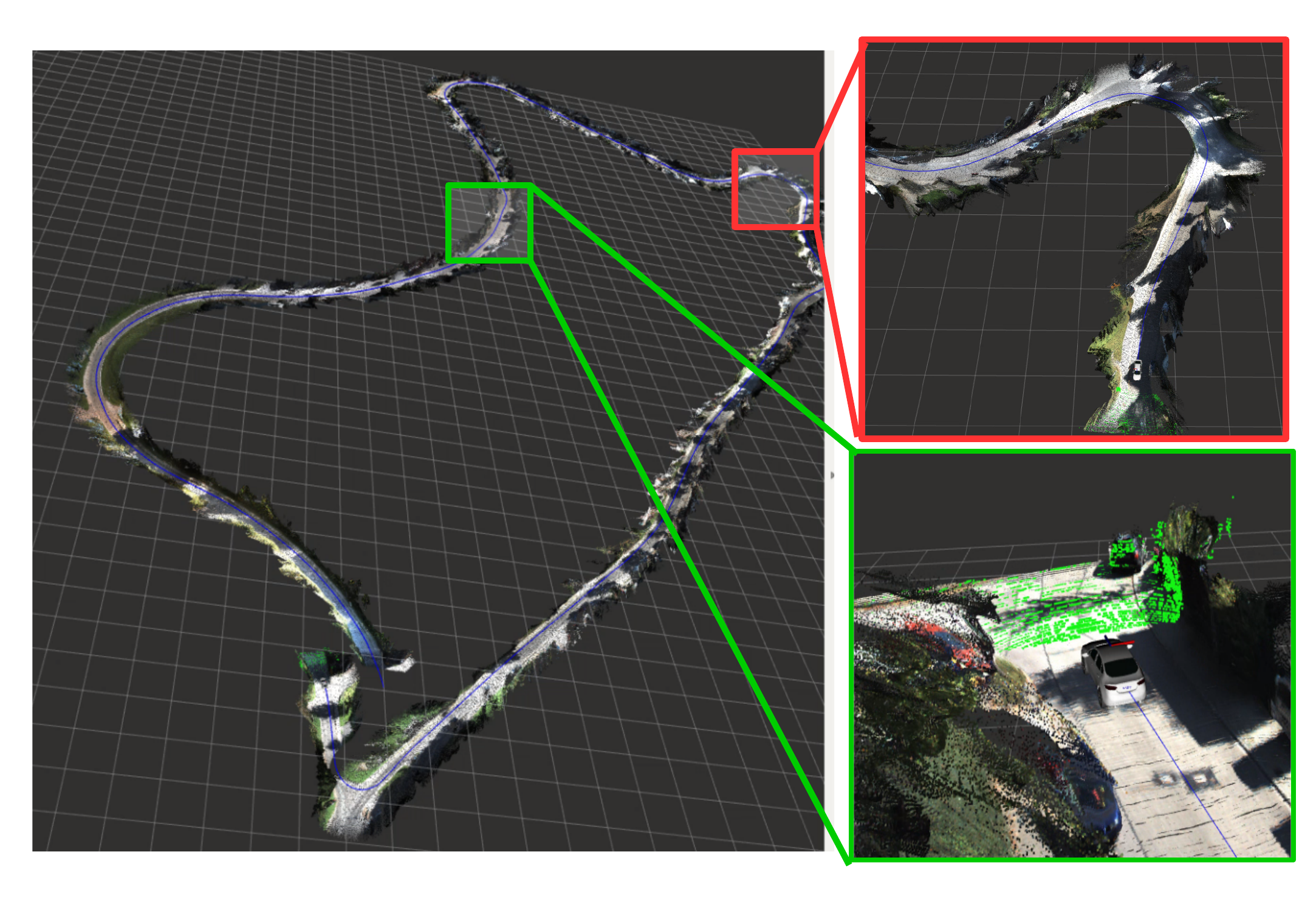}} \caption{Mapping
		results for KITTI Seq. 09 via our method. 3D points are first obtained
		by inverse projection with depth and camera intrinsics and then
		transformed into a global coordinate via the camera pose. The green
		points represent LiDAR points just for comparison. The blue curves
		describe the trajectory of our method.}
	\label{fig:head}
\end{figure}
\begin{figure*}[thpb]
	\centerline{\includegraphics[width=190mm]{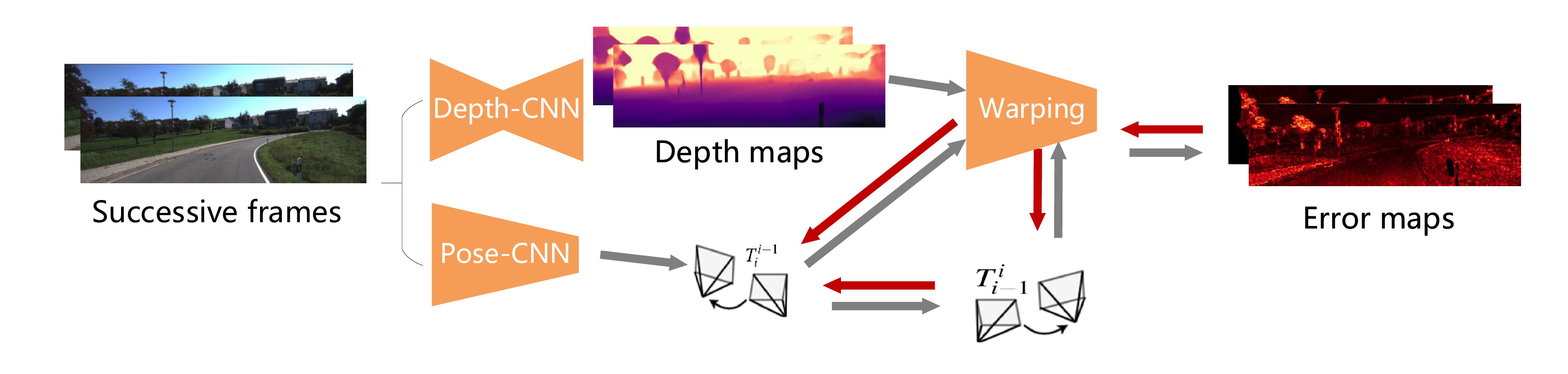}} \caption{The
	inference pipeline of our method for monocular VO. (1) A pair of consecutive
	frames are fed into the Depth-CNN and Pose-CNN to get depth maps and pose
	initialization. (2) Photometric error maps are calculated by forward and
	backward warping. (3) Relative transform $T_i^{i-1}$ are optimized by
	minimizing photometric errors. Gray arrows represent the error calculation
	path. Red arrows refer to gradient back-propagation.}
	\label{pipeline}
\end{figure*}

Traditional monocular VO methods \cite{ptam, ORB2, ORB3, svo, lsdslam, DSO}
usually consist of tracking, local or global optimization and loop closure
modules, which make full use of geometric constraints. Although some traditional
methods have shown excellent performance in terms of robustness and accuracy,
they inherently suffer from the loss of absolute scales without extra
information. Moreover, no reliable ego-motion will be obtained unless the
parallax between successive frames is large enough.

Deep learning based methods \cite{firstDVO, bian2019unsupervised, DeepVO, DPC,
	onlineadaption} try to tackle the above problems by training CNNs on large
amounts of data. Rather than specifying these geometric constraints manually,
DL-based methods are able to obtain them by exploiting prior knowledge from
training data. As a result, reasonable poses and depth can be estimated even
when parallax is not large enough. Besides, online-learning can be utilized to
further improve the performance. Despite the advantages of the DL-based methods,
the accuracy of estimated ego-motion is still inferior to traditional
approaches.


Hybrid methods try to combine the interpretability of traditional methods and
strong data fitting abilities of DL-based methods. Existing hybrid methods
usually leverage CNNs as initialization for traditional VO frameworks
\cite{cnnsvo, dvso, d3vo}. Though hybrid methods achieve state-of-the-art
results, the heavy calculation burdens prevent further application into
practice.

In this paper, we propose a novel deep online correction (DOC) framework for
monocular VO, which is composed of two CNNs and the proposed online correction
module. The former provides depth estimation and initial poses, while the latter
further improves the accuracy and robustness of poses via gradient propagations.
Different from existing hybrid methods, the whole pipeline is concise and does
not involve traditional modules like global bundle adjustment or pose graph
optimization. Different from existing online-learning methods, the parameters of
depth and pose CNNs will not be updated during inference phases, which improves
the efficiency of real-time performance. This approach is much more effective
because it reduces dimensions of optimizable space from millions of parameters
to only 6DoF poses. In addition, it can be implemented in a two-frame or
three-frame manner, which is quite flexible.

The contributions of our works are as follows:
\begin{itemize}
	\item We propose a fully DL-based VO which is composed of CNNs and an online
	      correction module. The proposed method combines the advantages of both
	      traditional and DL-based methods.
	\item Another version named DOC+ is designed to further improve the VO
	      performance. Two kinds of implementations are provided to show
	      flexibility.
	\item Our approach achieves state-of-the-art accuracy among current
	      monocular methods. On the KITTI Odometry Seq. 09 (see Fig.
	      \ref{fig:head} for visualization), DOC and DOC+ achieve performance of
	      RTE=2.3\% and RTE=2.0\% respectively.
\end{itemize}

\begin{figure*}[htpb]
	\centerline{\includegraphics[width=180mm]{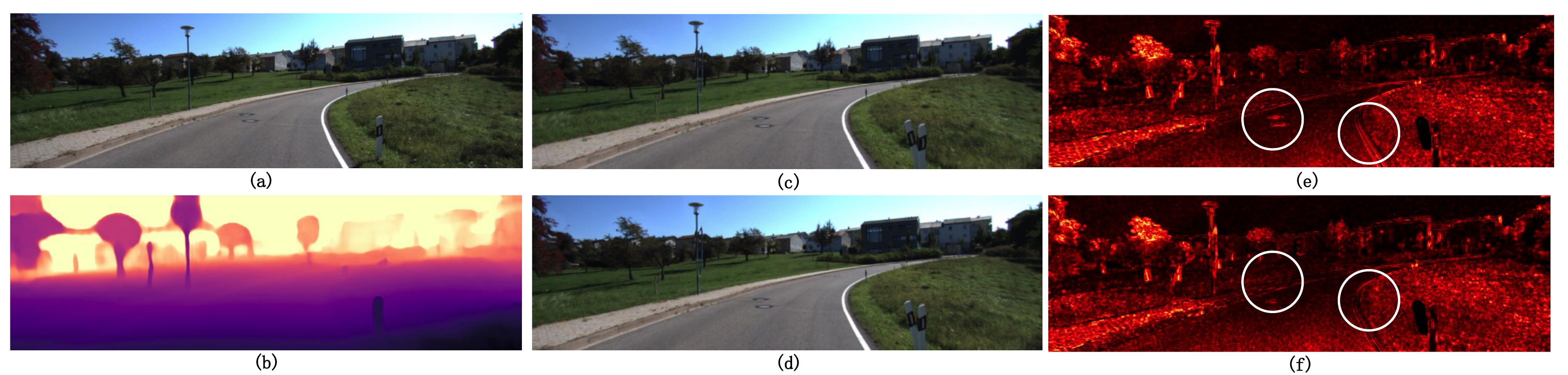}}
	\caption{Visualization of photometric errors before and after online
		correction. Sub-figure (a) and (b) are original images and corresponding
		depth maps respectively. Dark to bright in the depth maps represent near to
		far. Sub-figure (c) and(d) are reconstructed images before and after online
		correction. Sub-figure (e) and (f) are photometric errors before and after
		online correction. From dark to red represents errors from small to large.
		Note that in the error maps, errors around the curb and the manhole cover
		(circled out in the error map) have been reduced during the online
		correction.}
	\label{error_map}
\end{figure*}

\section{Related Works}
\label{sec:related}

\subsection{Traditional VO}

Traditional VO is a vital component for most popular SLAM frameworks. They can
be roughly divided into two categories: indirect methods \cite{ptam, ORB2, ORB3} that
recover depth and poses by minimizing geometric errors; direct methods
\cite{svo, DSO,lsdslam} that minimize photometric errors. Despite great accuracy
and efficiency, these methods usually can only estimate poses and depth up to an
unknown scale factor in a monocular setup. Moreover, accurate poses may not be
recovered in extreme conditions such as texture-less surfaces or dynamic scenes.

\subsection{Deep-learning-based VO}
Recent years have witnessed thrived development of deep learning based visual
odometry. One of the first DL-based VO estimation methods is proposed by Konda
et al. \cite{firstDVO} and Wang et al. \cite{DeepVO}. Despite impressive
results, their applications are limited by the requirements of labeled data.

To solve this problem, unsupervised methods are proposed and become popular.
SfM-Learner \cite{sfmlearner} proposed by Zhou et al. is a representative
pioneer of unsupervised VO. It contains a Pose-CNN and a Depth-CNN, and the
generated poses and depth are utilized to synthesize a new view as supervision.
However, the performance of SfM-Learner is less comparative to traditional
methods and it still suffers from scale ambiguities problem. Li et al.
\cite{undeepvo} and Bian et al. \cite{bian2019unsupervised} solve the scale
problem by introducing either stereo image training or extra scale consistent
loss term. The performance of unsupervised methods is further improved by some
following works: Monodepth2 \cite{monodepth2} utilize a minimum reprojection
loss and a multi-scale strategy during training for better depth estimation;
Zhao, W et al. \cite{zhao2020towards} proposed a novel framework that utilizes
optical flow to estimate poses; Wagstaff et al. \cite{DPC} proposed a two-stage
method in Deep Pose Correction (DPC) to further improve the results from
Pose-CNN. Since unsupervised learning does not rely on labeled data,
online-learning can be used to improve performance on test data: Li et al.
\cite{onlineadaption} utilize meta-learning for better generalizability. Chen et
al. \cite{GLNet} proposed the concept of the Output Fine Tune (OFT) and the
Parameter Fine Tune (PFT). However, the comparison has only been made for depth
refinements and the OFT shows little improvements to the overall results. While
our methods exploit the potential of OFT in visual odometry by designing
specific optimization procedure for online correction. Despite promising results
on visual odometry, DL-based methods are still inferior to some traditional
frameworks concerning generalizability and efficiency. We kindly recommend the
visual odometry survey \cite{surveyVO} for more details.

\subsection{Hybrid VO}
Since the pros and cons of traditional VO and DL-based VO vary, it is a natural
idea to combine them to build a robust VO system. CNN-SVO \cite{cnnsvo} is one
of the first hybrid VO frameworks by integrate a depth-CNN into SVO \cite{svo}.
DVSO \cite{dvso} and D3VO \cite{d3vo} use CNNs as initialization and use DSO
\cite{DSO} as optimization backend. Benefits from good initialization of CNNs
and strong robustness from traditional frameworks, hybrid methods have now
achieved the state-of-the-art result in public benchmark. Deep frame-to-frame
visual odometry (DFVO) \cite{DFVO} utilizes CNNs to predict optical flow and
depth of given image and then use traditional geometric constraints to recover
6DoF and achieve a promising result. BANet \cite{banet} modified the bundle
adjustment such that it is differentiable and thus can be integrated into an
end-to-end framework. Our approach is different from existing hybrid methods as
we propose an novel online correction module based on DL frameworks without
using global bundle adjustment.


\section{Approach}
\label{sec:approach}
\begin{figure}[thpb]
	\centerline{\includegraphics[width=90mm]{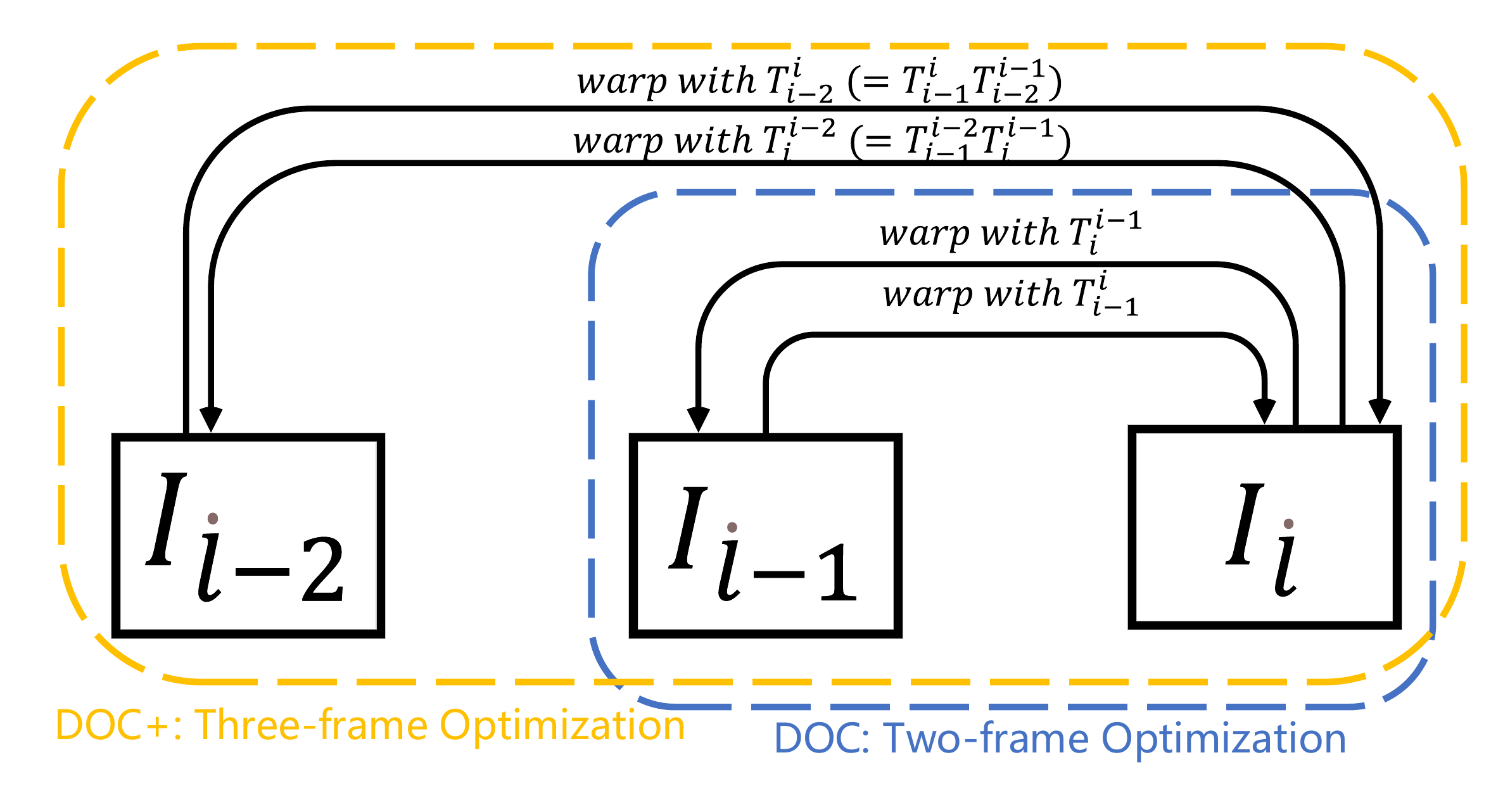}}
	\caption{Illustrations of DOC (two-frame) and DOC+ (three-frame)
		frameworks. The blue box represents DOC which only minimizes
		photometric errors from two consecutive frames. The orange box
		represents DOC+ which uses reprojection errors from pairs of frames
		among three frames.}
	\label{warping}
\end{figure}

The core idea of the DOC framework is that the relative poses are directly
optimized by minimizing photometric errors based on gradient propagation. DOC
does not rely on traditional frameworks and only need to calculate gradient
w.r.t the 6DoF pose (see Fig. \ref{pipeline} for details). In this section, we
first describe the training procedure for Depth-CNN and Pose-CNN. And then we
describe details of DOC (two-frame optimization) and DOC+ (three-frame
optimization) as shown in Fig \ref{warping}.

\subsection{Training of Depth-CNN and Pose-CNN}
For training, the proposed Depth-CNN and Pose-CNN have similar architectures to
Monodepth2 \cite{monodepth2}, but are tailored for the need for online
correction initialization. Depth-CNN has a U-Net like structure with
skip-connections. It takes a single RGB image and outputs a depth map. Pose-CNN
takes two concatenated images as input and outputs rotation and translation
vectors. Stereo images are leveraged in training phases to recover absolute
scales, while only monocular images are fed into the networks during testing.
Besides, to improve the performance of the networks and for the usage of online
correction model. We use the ``explainability" mask \cite{sfmlearner} instead of
auto-masking in original paper.

\subsection{DOC: Two-frame-based Optimization}
For testing, DOC considers two consecutive frames in the online correction
module. Given the input images \{$I_{i-1}$, $I_{i}$\}, the Depth-CNN and
Pose-CNN are used to the infer depth \{$D_{i-1}$, $D_{i}$\} and initial pose
$T_{i}^{i-1}$ respectively. The online correction module will iteratively
refine the ego-motion via minimizing special photometric errors defined as
\begin{equation}
	E_i = E_{i-1}^{i} + E_{i}^{i-1},
\end{equation}
where $E_i$ is the photometric errors at time step i. $E_{i-1}^{i}$ and
$E_{i}^{i-1}$ are forward and backward errors as follows:
\begin{equation}
	E_{i-1}^{i}  = M_i \circ E_{pho}(\Pi(D_i, T_{i}^{i-1}, I_{i-1}, K), I_i), \label{loss1}
\end{equation}
\begin{equation}
	E_{i}^{i-1}  = M_{i-1} \circ E_{pho}(\Pi(D_{i-1}, T_{i-1}^{i}, I_{i}, K), I_{i-1}), \label{loss2}
\end{equation}
where $\circ$ denotes element-wise multiplication. Here we take $E_{i-1}^{i}$
for an illustration and $E_{i}^{i-1}$ can be obtained in a similar way.

$\Pi$ in (\ref{loss1}) is the warping function as proposed by Jaderberg et al.
in \cite{stn}. It synthesis a novel view according to input image $I_i$,
corresponding depth $D_i$, camera intrinsic $K$ and transform matrix
$T_{i}^{i-1}$ by
\begin{equation}
	I_i^{'} = \Pi(D_i, T_{i}^{i-1}, I_{i-1}, K). \label{warp}
\end{equation}
$E_{pho}$ in (\ref{loss1}) calculate photometric errors between images:
\begin{equation}
	E_{pho}(I_{i}, I_{i-1}) = M_{std}(||I_{i} - I_{i-1}||_1), \label{pho}
\end{equation}
where $M_{std}$ is a function that behaves as an outlier rejecter:
\begin{equation}
	M_{std}(E) = E \circ (E < (\bar{E} + E_{\sigma})), \label{mstd}
\end{equation}
where $\bar{E}$ and $E_{\sigma}$ are means and standard deviations of $E$
respectively. Instead of using SSIM \cite{ssim}, we use the truncated $L_1$
loss because we find it achieves similar accuracy for the DOC module but is more
computational efficient.

$M_i$ in (\ref{loss1}) is a mask composed of an occlusion mask and an
explainability mask denoted as $M_o$ and $M_e$ respectively, $M$ = $M_o \circ
	M_e$. Similar to \cite{inthewild}, the occlusion mask $M_o$ is defined as
follows:
\begin{equation}
	M_o = [\Pi(D_i, \hat{T_{i}}, D_{i-1}, K) > D_i | D_i > d_m], \label{occmask}
\end{equation}
where $[\bullet]$ is the Iverson bracket and $d_m$ is a threshold set to 5
meters to filter out wrong occlusion mask pixels caused by incorrect depth
estimation especially in the distance. $M_e$ is the explainability mask
predicted by CNN according to SfM-Learner \cite{sfmlearner}. As we can see from
Fig.~\ref{maskvis}, $M_o$ successfully covers the pixels with ``ghosting effect"
where the double traffic rods appear. $M_e$ reduces photometric errors in the
high-frequency areas like roofs and vegetation.

The equation (\ref{loss1}) and (\ref{loss2}) is connected with the following
constraint:
\begin{equation}
	T_{i-1}^{i} = inv(T_{i}^{i-1}) \label{invT}.
\end{equation}

$T_{i-1}^{i}$ represents a rigid transform $\in \mathbb{R}^{4x4}$ parameterized
by a 6 degree vector $r_{i}^{i-1}, t_{i}^{i-1} \in SE(3)$. The vector can be
easily transformed into transform matrix with the Rodrigues' rotation formula.

Finally, Adam \cite{adam} optimizer is used to update relative poses. The
maximum number of iterations $N$ is set to 20 in all our experiments. The full
algorithm is presented in Alg. \ref{pseuducode}. The visualization and
discussion of error maps before and after optimization are shown in Fig.
\ref{error_map}.

\subsection{DOC+: Three-frame-based Optimization}
DOC+ takes three frames in the online correction module, which can further
improve the pose accuracy.

For each frame $I_i$, we consider two previous frames $I_{i-1}$ and $I_{i-2}$.
The relative pose for last frames $T_{i-1}^{i-2}$ has been optimized during last
step. $T_{i}^{i-1}$ is initialized by Pose-CNN and will be further optimized. We
consider 4 photometric errors into the total energy function:
\begin{equation}
	E_i = \alpha (E_{i-1}^{i} + E_{i}^{i-1}) +
	(1-\alpha) (E_{i}^{i-2} + E_{i-2}^{i}),
\end{equation}
where $\alpha$ is a balancing factor between current and previous frames and is
set to 0.8 during all experiments.

\begin{algorithm}
	\caption{Deep Online Correction (two-frame based)}
	\label{pseuducode}
	\begin{flushleft}
		\textbf{Require:} Depth-CNN; Pose-CNN; Intrinsic: $K$\\
		\textbf{Input:} Image sequence: [$I_1, I_2, ...,I_k$]\\
		\textbf{Output:} Refined Pose: [$T_1, T_2, ...,T_k$]
	\end{flushleft}
	\begin{algorithmic}
		\State \textbf{Initialization:} $T_1 = I$
		\For {$i=2,3, \ldots,k$}
		\State Get Depth-CNN Prediction $D_{i-1}, D_{i}$
		\State Get Pose-CNN Prediction $\hat{r_i}^{i-1}, \hat{t_i}^{i-1}$
		\State Compute transform matrix $\hat{T_{i}}^{i-1}$ from $\hat{r_i}^{i-1}, \hat{t_i}^{i-1}$
		\For {$step=1, 2, \ldots, N$}
		\State Warp $I_{i-1}$ into $I_i$: $I_{i}^{'} = \Pi(D_i, \hat{T_{i}}^{i-1}, I_{i-1}, K)$
		\State Compute $E_{i-1}^{i}$ by (\ref{loss1})
		\State Warp $I_{i}$ into $I_{i-1}$: $I_{i-1}^{'} = \Pi(D_{i-1}, \hat{T}_{i-1}^{i}, I_{i}, K)$
		\State Compute $E_{i}^{i-1}$ by (\ref{loss2})
		\State $E_i = E_{i-1}^{i} + E_{i}^{i-1}$
		\State Compute gradient w.r.t. $\hat{r_i}^{i-1}, \hat{t_i}^{i-1}$
		\State Use Adam optimizer to update $\hat{r_i}^{i-1}, \hat{t_i}^{i-1}$
		\EndFor
		\State $T_i \leftarrow T_{i-1} T_{i}^{i+1} $
		\EndFor
	\end{algorithmic}
\end{algorithm}

The photometric errors is defined similar as (\ref{loss1}) and (\ref{loss2}).
Once $T_{i}^{i-1}$ is obtained, $T_{i}^{i-2}$ will be calculated as
\begin{equation}
	T_{i}^{i-2} = T_{i-1}^{i-2}  T_{i}^{i-1}.
\end{equation}

The Adam optimizer is used to minimize $E_i$ with respect to $T_{i-1}^{i-2}$ and
$T_{i}^{i-1}$. Since the former one has been already updated during the last
optimization, it is a natural idea to prevent $T_{i-1}^{i-2}$ from being updated
too much from initial values. In traditional frameworks, this is usually
achieved by marginalization. We achieve this by setting different learning rates
for each step. We empirically set the learning rate of previous frames as 10
times less than current frames.


\section{Experiments}
\label{sec:experiments}

\subsection{Implementation Details}

We conduct experiments on both KITTI odometry dataset \cite{Kitti1, Kitti2} and
EuRoC MAV dataset \cite{euroc}. For KITTI odometry dataset, the input images are
resized to 832$\times$256. The Pose-CNN and Depth-CNN are jointly trained on
sequence 00-08, which contains 36671 training frames in total, and then
validated on sequence 09 and 10 combined with the online correction module. The
relative translation error (RTE) and relative rotation error (RRE) are applied
for evaluation. RTE is the average translational root mean square error (RMSE)
drift in percentage on length from 100, 200, ..., 800m, while RRE is the average
rotation RMSE drift (\degree/100m) on length from 100, 200, ..., 800m. It is
worth noting that we do not use the full KITTI Eigen split for training since it
has some overlaps with KITTI odometry test dataset. For EuRoC MAV dataset, all
stereo images are rectified and then resized to 736$\times$480. Sequence MH\_03
and MH\_05 are used for testing. All the other sequences, which contains 22067
frames, are used for training. The RMSE of absolute trajectory errors (ATE) is
used as evaluation metrics.

\begin{figure}[thpb]
	\centerline{\includegraphics[width=80mm]{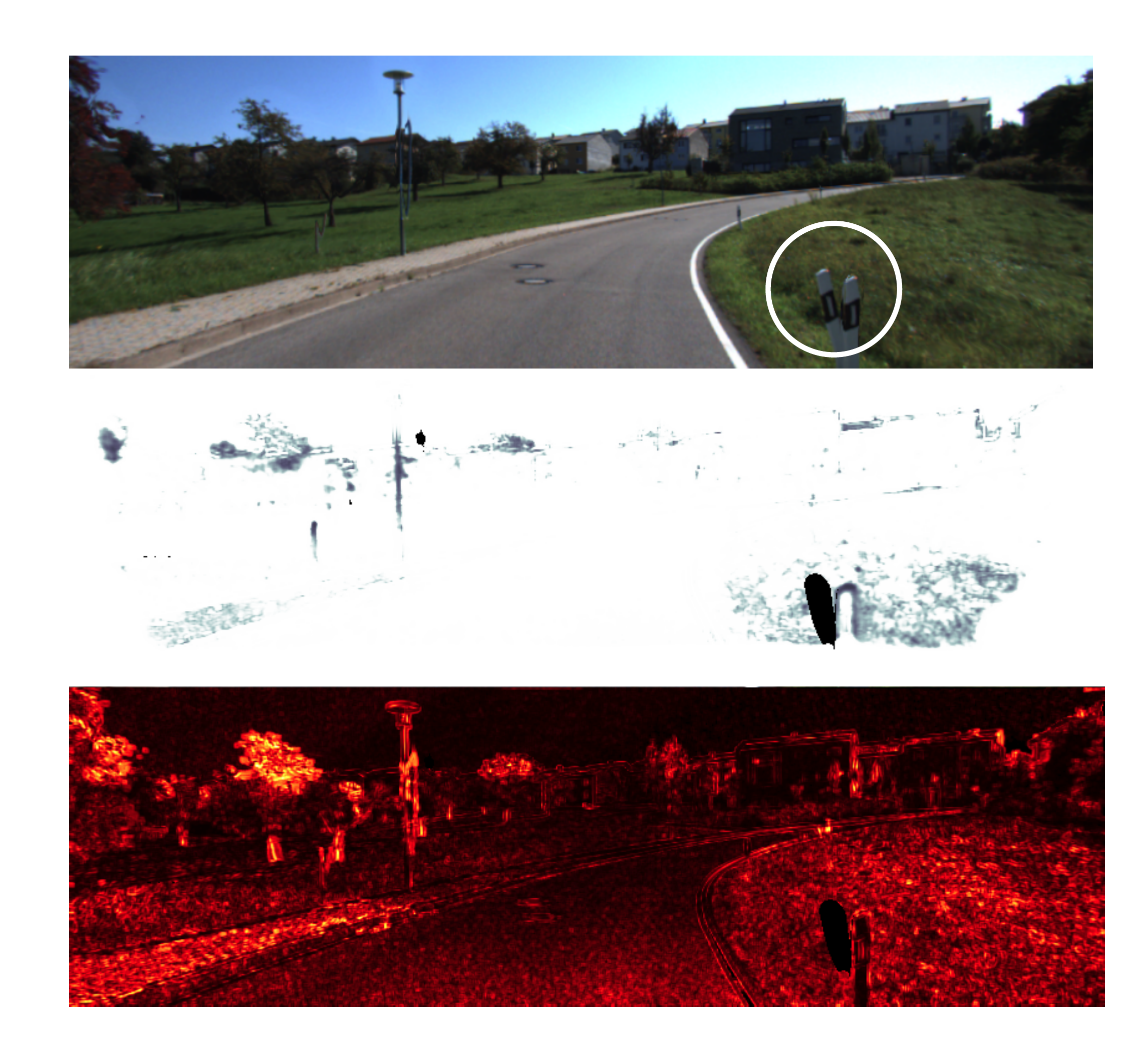}} \caption{Visualization.
		From top to bottom: warped image, the combination of occlusion (black)
		and explainability (gray) masks, photometric error map. In the top, we
		can see the double traffic rods in the warped image produced by
		occlusion during backward warping. The occlusion mask successfully
		calculates the area of pixels where occlusion happens. As a result the
		occluded region is not been calculated in photometric errors. The
		explainability mask in the middle produced by CNN usually reduces
		photometric errors in the high-frequency areas like roofs and
		vegetation.}
	\label{maskvis}
\end{figure}

\begin{figure}[thpb]
	\centering
	\begin{subfigure}[b]{0.5\textwidth}
		\centering
		\includegraphics[width=\textwidth]{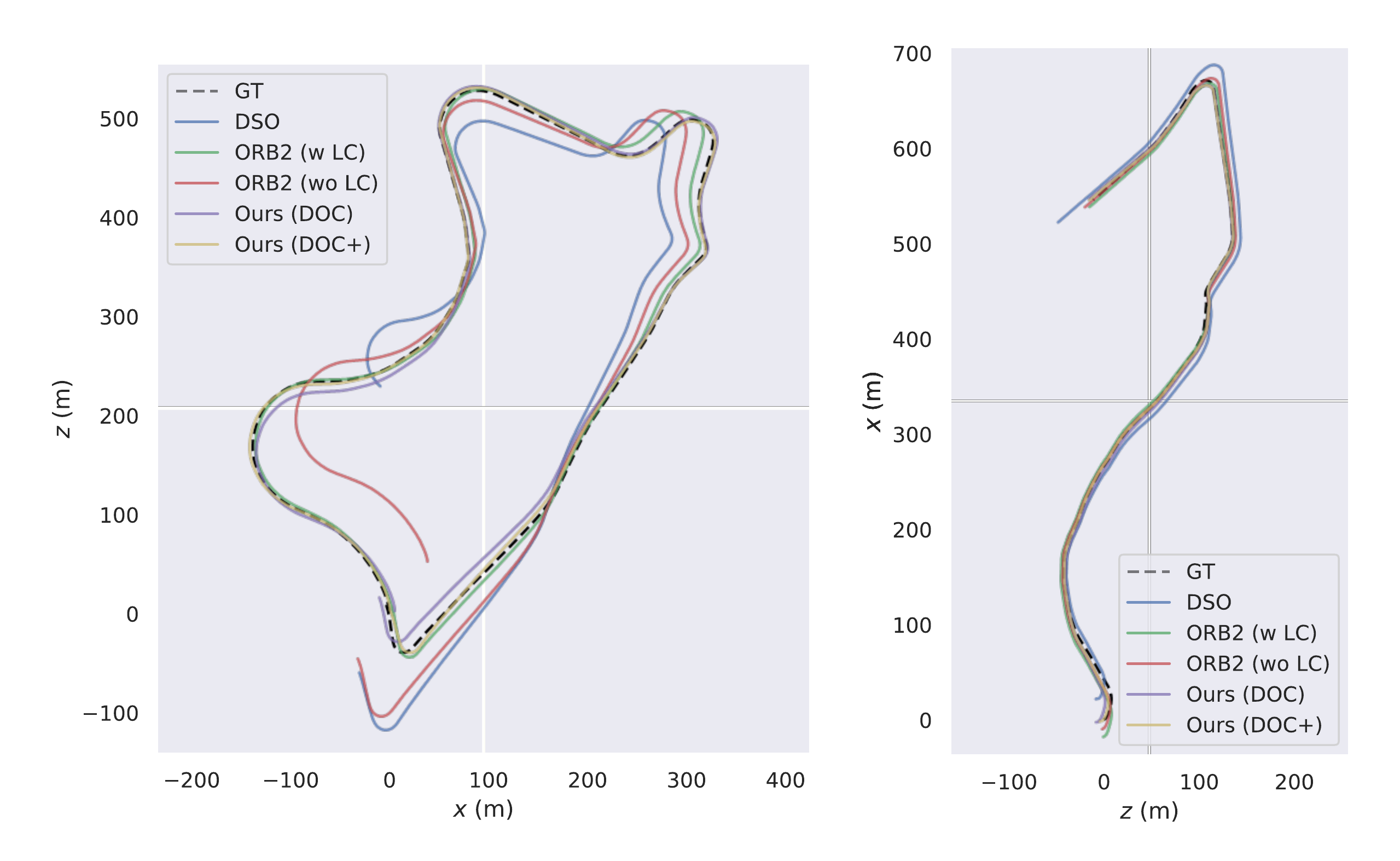}
		\caption{Trajectories of our methods and traditional methods.}
		\label{fig_traj_traditional}
	\end{subfigure}
	\hfill
	\begin{subfigure}[b]{0.5\textwidth}
		\centering
		\includegraphics[width=\textwidth]{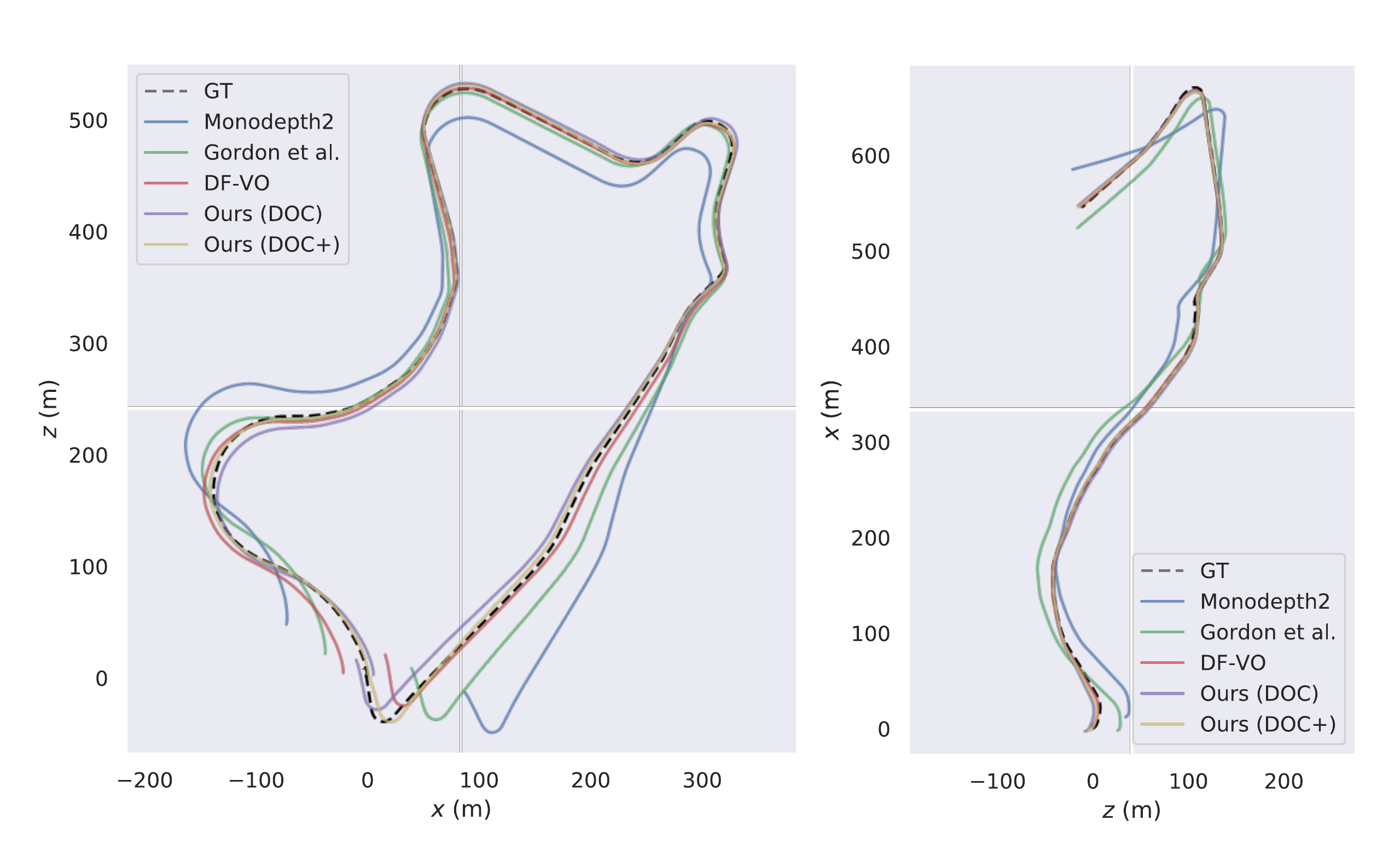}
		\caption{Trajectories of our methods, DL-based and hybrid methods.}
		\label{fig_traj_hybrid}
	\end{subfigure}
	\caption{Comparison of our methods with traditional methods and hybrid
		methods. Comparative experiments are conducted on KITTI Odometry Seq. 09
		(left) and Seq. 10 (right). Figures in (a) show results of our method
		and traditional methods, while figures in (b) demonstrate the
		trajectories of our methods and hybrid methods.}
	\label{fig_traj}
\end{figure}

For both datasets, stereo images are utilized to train the CNNs while only
monocular images are needed in the test phases. With the prior knowledge learned
from stereo images, our method can recover the absolute scales even for unseen
images. Backbones of Depth-CNN and Pose-CNN are ResNet-18 \cite{resnet} with
pretrained weights from ImageNet \cite{imagenet}. We train the Depth-CNN and
Pose-CNN for 20 epochs and use the parameters from last epoch to get depth
estimation and pose initialization for testing sequences. The Adam optimizer
with a learning rate of 1e-4 for the first 15 epochs and 1e-5 for the last 5
epochs are used during training. The batch size is set to 8 for training and 1
for online correction module. The whole framework is implemented by PyTorch
\cite{pytorch} on a single NVIDIA TITAN Xp GPU. To speed up the online
correction process, the whole online correction is slightly modified from Alg.
\ref{pseuducode}. The equation \ref{warp} is composed of three parts:
unprojection, transform and projection. For every frame, the unprojection part
does not involve gradient propagation. Thus, it can be pre-computed before
optimization iterations. The running speed for DOC and DOC+ is about 8 FPS and 5
FPS respectively. The running time includes both CNN inference and online
correction modules.

\begin{table*}[thpb]
	\begin{center}
		\begin{tabular}{|c||c|c|c|c|c|c|}
			\hline
			                                        & \multicolumn{3}{c|}{Sequence 09} & \multicolumn{3}{c|}{Sequence 10}                                                                      \\\hline
			Methods                                 & RTE (\%)                         & RRE (\degree/100m)               & ATE (m)       & RTE (\%)      & RRE (\degree/100m) & ATE (m)       \\\hline
			DSO \cite{DSO}                          & 15.91                            & \textbf{0.20}                    & 52.23         & 6.49          & \textbf{0.20}      & 11.09         \\\hline
			ORB-SLAM2 (w/o LC) \cite{ORB2}          & 9.3                              & 0.26                             & 38.77         & 2.57          & 0.32               & 5.42          \\\hline
			ORB-SLAM2 (w/ LC) \cite{ORB2}           & 2.88                             & 0.25                             & 8.39          & 3.30          & 0.30               & 6.63          \\\hline
			SC-SfM-Learner \cite{scsfmlearner}      & 7.64                             & 2.19                             & 15.02         & 10.74         & 4.58               & 20.19         \\\hline
			Monodepth2 \cite{monodepth2}            & 14.89                            & 3.44                             & 68.75         & 11.29         & 4.97               & 21.93         \\\hline
			Online Adaptation \cite{onlineadaption} & 5.89                             & 3.34                             & -             & 4.79          & 0.83               & -             \\\hline
			DPC \cite{DPC}                          & 2.82                             & 0.76                             & -             & 3.81          & 1.34               & -             \\\hline
			DFVO (Stereo Trained) \cite{DFVO}       & 2.61                             & 0.29                             & 10.88         & \textbf{2.29} & 0.37               & 3.72          \\\hline
			Gordon et al. \cite{inthewild}          & 2.7                              & -                                & -             & 6.8           & -                  & -             \\\hline
			Ours (DOC)                              & 2.26                             & 0.87                             & 7.34          & 2.61          & 1.59               & 4.23          \\\hline
			Ours (DOC+)                             & \textbf{2.02}                    & 0.61                             & \textbf{4.76} & \textbf{2.29} & 1.10               & \textbf{3.38} \\\hline
		\end{tabular}
	\end{center}
	\caption{Monocular visual odometry comparison on KITTI Odometry Seq.09 and
		Seq.10, with different approaches including traditional, DL-based and
		hybrid methods. RTE, RRE and ATE are abbreviations for relative
		translation errors, relative rotation errors and absolute translation
		errors respectively.}
	\label{comparison}
\end{table*}

\begin{table*}[thpb]
	\begin{center}
		\begin{tabular}{|c|c||c|c|c|c|c||c|c|c|}
			\hline
			    & Method                 & $M_e$      & $M_o$      & Loss                 & Frames & RTE (\%) & RRE (\degree/100m) & ATE (m) \\\hline
			(a) & DOC w/o mask           &            &            & $L_1$ w/ truncation  & 2      & 2.66     & 1.00               & 12.92   \\\hline
			(b) & w/ explainability mask & \checkmark &            & $L_1$ w/ truncation  & 2      & 2.53     & 0.93               & 12.54   \\\hline
			(c) & w/ occlusion mask      &            & \checkmark & $L_1$ w/ truncation  & 2      & 2.54     & 0.96               & 11.42   \\\hline
			(d) & DOC w/o truncation     & \checkmark & \checkmark & $L_1$ w/o truncation & 2      & 2.81     & 1.07               & 13.30   \\\hline
			(e) & DOC w/ SSIM            & \checkmark & \checkmark & SSIM                 & 2      & 2.30     & 0.82               & 7.31    \\\hline
			(f) & DOC                    & \checkmark & \checkmark & $L_1$ w/ truncation  & 2      & 2.26     & 0.87               & 7.34    \\\hline
			(g) & DOC+ w/ SSIM           & \checkmark & \checkmark & SSIM                 & 3      & 2.03     & 0.68               & 5.51    \\\hline
			(h) & DOC+                   & \checkmark & \checkmark & $L_1$ w/ truncation  & 3      & 2.02     & 0.61               & 4.76    \\\hline
		\end{tabular}
	\end{center}
	\caption{Ablation results are evaluated on KITTI Odometry Seq. 09. $M_e$ and
		$M_o$ refer to explainability mask and occlusion mask respectively.
		``Loss" means which loss to
		use during online correction methods. ``Frames" = 2 or 3 stands for
		two-frame or three-frame optimization.}
	\label{ablation}
\end{table*}



\subsection{Visual Odometry Evaluation}
For KITTI odometry dataset, we take Pose-CNN from Monodepth2 as our baseline. As
illustrated in Table \ref{comparison}, our proposed DOC method outperforms
traditional and DL-based methods and is competitive to existing hybrid methods.
Trajectories of these methods are visualized in Fig. \ref{fig_traj} using EVO
\cite{evo}. Li et al. \cite{onlineadaption} use the meta-learning technique to
update parameters during test time. Compared to our approach, we only need to
calculate gradient propagation to 6DoF poses without updating the whole
networks. DFVO \cite{DFVO} achieves better results in RRE as it uses traditional
essential matrix estimation and PnP method to recover pose from depth and flow.
As we can see from Seq. 09, traditional methods usually suffer from scale drift
problems in monocular setups without loop closure. While DL-based methods can
not guarantee accurate poses throughout the whole trajectories. It is observable
that even with inferior performance on RTE, traditional methods usually show
better performance on RRE than DL-based methods. The reason behind maybe
twofold: First, scale drift problem clearly does not affect RRE as rotation is
scale invariant. Second, traditional methods explicitly model optimization in
rotation manifold SO(3) while DL-based methods numerically solve these problems
through gradient update and Adam optimizer. Compared to both traditional and
existing DL-based methods, our methods (DOC and DOC+) clearly show better
performance and is comparable to hybrid methods. Besides, the trajectories
estimated by DOC+ clearly show very small translational drift (with lowest ATE
among all methods) without loop closure.

For EuRoC MAV dataset, we use ATE as evaluation metrics. It is a very
challenging dataset as it contains large motion and various illumination
conditions. As shown in Table \ref{eurocresult}, the proposed methods clearly
improve the odometry accuracy from Monodepth2 initialization. Our methods also
outperform traditional methods like DSO and is comparable with ORB-SLAM. Indoor
datasets like EuRoC MAV usually have trajectories in one room or small space.
Thus a full SLAM system with re-localization like ORB-SLAM usually perform
better than pure VO methods.

\begin{table}[thpb]
	\begin{center}
		\begin{tabular}{|c||c|c|}
			\hline
			Methods                       & MH\_03        & MH\_05        \\\hline
			DSO \cite{DSO}                & 0.17          & 0.11          \\\hline
			ORB-SLAM  \cite{ORB1}         & \textbf{0.08} & 0.16          \\\hline
			Monodepth2  \cite{monodepth2} & 1.88          & 1.56          \\\hline
			Ours (DOC)                    & 0.15          & 0.11          \\\hline
			Ours (DOC+)                   & 0.13          & \textbf{0.09} \\\hline
		\end{tabular}
	\end{center}
	\caption{Monocular visual odometry comparison on EuRoC MAV dataset. The RMSE
		of absolute trajectory errors (ATE) is used as evaluation metrics.}
	\label{eurocresult}
\end{table}

\subsection{Ablation Study}
We conduct a detailed ablation study for the DOC module on KITTI datasets (see
Table \ref{ablation}). First, we explore different loss functions for
photometric errors. We found that truncated L1 loss can achieve similar
performance compared to SSIM loss (e, f, g, h). The calculation of SSIM loss
requires the computation of local mean and variance for every pixel while L1
loss between two images is clearly more computationally efficient. Second, we
found that both explainability masks and occlusion masks improve the overall
result (a-c). Occlusion masks stop gradient propagation where pixels have
``ghosting effect" (see Fig. \ref{maskvis}). Explainability masks produced by
Depth-CNN reduce weights for pixels usually occupied by high-frequency areas
like trees and roofs or non-Lambertian surfaces like windows. These areas are
noises for online correction and can not be described by reprojection warping.
Finally, with all methods mentioned above, by only using the consecutive frame
for online correction, our method already achieve satisfying odometry result. By
using the three-frame-based optimization, DOC+ further achieve better result and
smaller translation drift thanks to reprojection constraints from more image
pairs. Adding more frames for optimization shows very little improvements for
accuracy while increases the overall running time.

\section{CONCLUSIONS}
\label{sec:conclusions}

In this paper, we propose a novel monocular visual odometry algorithm with an
online correction module. It relies on DL-based frameworks and leverages
advantages of both CNNs and geometric constraints. Specifically, Depth-CNN and
Pose-CNN are trained in a self-supervised manner to provided initial ego-motion
and depth maps with absolute scales. And then a novel online correction module
based on gradient back-propagation is proposed to further improve the VO
accuracy. Different from existing online learning methods, our online correction
module does not update the networks' parameters which makes it more concise and
computationally efficient. Experiment results on KITTI dataset demonstrate that
our method outperforms existing traditional and DL-based methods and is
comparable with state-of-the-art hybrid methods.

\addtolength{\textheight}{-3.3cm}   








\bibliographystyle{IEEEtran}
\bibliography{root}

\begin{thebibliography}{10}
\providecommand{\url}[1]{#1}
\csname url@rmstyle\endcsname
\providecommand{\newblock}{\relax}
\providecommand{\bibinfo}[2]{#2}
\providecommand\BIBentrySTDinterwordspacing{\spaceskip=0pt\relax}
\providecommand\BIBentryALTinterwordstretchfactor{4}
\providecommand\BIBentryALTinterwordspacing{\spaceskip=\fontdimen2\font plus
\BIBentryALTinterwordstretchfactor\fontdimen3\font minus
  \fontdimen4\font\relax}
\providecommand\BIBforeignlanguage[2]{{%
\expandafter\ifx\csname l@#1\endcsname\relax
\typeout{** WARNING: IEEEtran.bst: No hyphenation pattern has been}%
\typeout{** loaded for the language `#1'. Using the pattern for}%
\typeout{** the default language instead.}%
\else
\language=\csname l@#1\endcsname
\fi
#2}}

\bibitem{tutorialVO}
D.~Scaramuzza and F.~Fraundorfer, ``Visual odometry [tutorial],'' \emph{IEEE
  robotics \& automation magazine}, vol.~18, no.~4, pp. 80--92, 2011.

\bibitem{ptam}
G.~Klein and D.~Murray, ``Parallel tracking and mapping for small ar
  workspaces,'' in \emph{2007 6th IEEE and ACM international symposium on mixed
  and augmented reality}.\hskip 1em plus 0.5em minus 0.4em\relax IEEE, 2007,
  pp. 225--234.

\bibitem{ORB2}
R.~Mur-Artal and J.~D. Tard{\'o}s, ``Orb-slam2: An open-source slam system for
  monocular, stereo, and rgb-d cameras,'' \emph{IEEE Transactions on Robotics},
  vol.~33, no.~5, pp. 1255--1262, 2017.

\bibitem{ORB3}
C.~Campos, R.~Elvira, J.~J.~G. Rodr{\'\i}guez, J.~M. Montiel, and J.~D.
  Tard{\'o}s, ``Orb-slam3: An accurate open-source library for visual,
  visual-inertial and multi-map slam,'' \emph{arXiv preprint arXiv:2007.11898},
  2020.

\bibitem{svo}
C.~Forster, M.~Pizzoli, and D.~Scaramuzza, ``Svo: Fast semi-direct monocular
  visual odometry,'' in \emph{2014 IEEE international conference on robotics
  and automation (ICRA)}.\hskip 1em plus 0.5em minus 0.4em\relax IEEE, 2014,
  pp. 15--22.

\bibitem{lsdslam}
J.~Engel, T.~Sch{\"o}ps, and D.~Cremers, ``Lsd-slam: Large-scale direct
  monocular slam,'' in \emph{European conference on computer vision}.\hskip 1em
  plus 0.5em minus 0.4em\relax Springer, 2014, pp. 834--849.

\bibitem{DSO}
J.~Engel, V.~Koltun, and D.~Cremers, ``Direct sparse odometry,'' \emph{IEEE
  transactions on pattern analysis and machine intelligence}, vol.~40, no.~3,
  pp. 611--625, 2017.

\bibitem{firstDVO}
K.~R. Konda and R.~Memisevic, ``Learning visual odometry with a convolutional
  network.'' in \emph{VISAPP (1)}, 2015, pp. 486--490.

\bibitem{bian2019unsupervised}
J.~Bian, Z.~Li, N.~Wang, H.~Zhan, C.~Shen, M.-M. Cheng, and I.~Reid,
  ``Unsupervised scale-consistent depth and ego-motion learning from monocular
  video,'' in \emph{Advances in neural information processing systems}, 2019,
  pp. 35--45.

\bibitem{DeepVO}
S.~Wang, R.~Clark, H.~Wen, and N.~Trigoni, ``Deepvo: Towards end-to-end visual
  odometry with deep recurrent convolutional neural networks,'' in \emph{2017
  IEEE International Conference on Robotics and Automation (ICRA)}.\hskip 1em
  plus 0.5em minus 0.4em\relax IEEE, 2017, pp. 2043--2050.

\bibitem{DPC}
B.~Wagstaff, V.~Peretroukhin, and J.~Kelly, ``Self-supervised deep pose
  corrections for robust visual odometry,'' \emph{arXiv preprint
  arXiv:2002.12339}, 2020.

\bibitem{onlineadaption}
S.~Li, X.~Wang, Y.~Cao, F.~Xue, Z.~Yan, and H.~Zha, ``Self-supervised deep
  visual odometry with online adaptation,'' in \emph{Proceedings of the
  IEEE/CVF Conference on Computer Vision and Pattern Recognition}, 2020, pp.
  6339--6348.

\bibitem{cnnsvo}
S.~Y. Loo, A.~J. Amiri, S.~Mashohor, S.~H. Tang, and H.~Zhang, ``Cnn-svo:
  Improving the mapping in semi-direct visual odometry using single-image depth
  prediction,'' in \emph{2019 International Conference on Robotics and
  Automation (ICRA)}.\hskip 1em plus 0.5em minus 0.4em\relax IEEE, 2019, pp.
  5218--5223.

\bibitem{dvso}
N.~Yang, R.~Wang, J.~Stuckler, and D.~Cremers, ``Deep virtual stereo odometry:
  Leveraging deep depth prediction for monocular direct sparse odometry,'' in
  \emph{Proceedings of the European Conference on Computer Vision (ECCV)},
  2018, pp. 817--833.

\bibitem{d3vo}
N.~Yang, L.~v. Stumberg, R.~Wang, and D.~Cremers, ``D3vo: Deep depth, deep pose
  and deep uncertainty for monocular visual odometry,'' in \emph{Proceedings of
  the IEEE/CVF Conference on Computer Vision and Pattern Recognition}, 2020,
  pp. 1281--1292.

\bibitem{sfmlearner}
T.~Zhou, M.~Brown, N.~Snavely, and D.~G. Lowe, ``Unsupervised learning of depth
  and ego-motion from video,'' in \emph{Proceedings of the IEEE Conference on
  Computer Vision and Pattern Recognition}, 2017, pp. 1851--1858.

\bibitem{undeepvo}
R.~Li, S.~Wang, Z.~Long, and D.~Gu, ``Undeepvo: Monocular visual odometry
  through unsupervised deep learning,'' in \emph{2018 IEEE international
  conference on robotics and automation (ICRA)}.\hskip 1em plus 0.5em minus
  0.4em\relax IEEE, 2018, pp. 7286--7291.

\bibitem{monodepth2}
C.~Godard, O.~Mac~Aodha, M.~Firman, and G.~J. Brostow, ``Digging into
  self-supervised monocular depth estimation,'' in \emph{Proceedings of the
  IEEE international conference on computer vision}, 2019, pp. 3828--3838.

\bibitem{zhao2020towards}
W.~Zhao, S.~Liu, Y.~Shu, and Y.-J. Liu, ``Towards better generalization: Joint
  depth-pose learning without posenet,'' in \emph{Proceedings of the IEEE/CVF
  Conference on Computer Vision and Pattern Recognition}, 2020, pp. 9151--9161.

\bibitem{GLNet}
Y.~Chen, C.~Schmid, and C.~Sminchisescu, ``Self-supervised learning with
  geometric constraints in monocular video: Connecting flow, depth, and
  camera,'' in \emph{Proceedings of the IEEE international conference on
  computer vision}, 2019, pp. 7063--7072.

\bibitem{surveyVO}
C.~Chen, B.~Wang, C.~X. Lu, N.~Trigoni, and A.~Markham, ``A survey on deep
  learning for localization and mapping: Towards the age of spatial machine
  intelligence,'' \emph{arXiv preprint arXiv:2006.12567}, 2020.

\bibitem{DFVO}
H.~Zhan, C.~S. Weerasekera, J.-W. Bian, and I.~Reid, ``Visual odometry
  revisited: What should be learnt?'' in \emph{2020 IEEE International
  Conference on Robotics and Automation (ICRA)}.\hskip 1em plus 0.5em minus
  0.4em\relax IEEE, 2020, pp. 4203--4210.

\bibitem{banet}
C.~Tang and P.~Tan, ``Ba-net: Dense bundle adjustment network,'' \emph{arXiv
  preprint arXiv:1806.04807}, 2018.

\bibitem{stn}
M.~Jaderberg, K.~Simonyan, A.~Zisserman, \emph{et~al.}, ``Spatial transformer
  networks,'' in \emph{Advances in neural information processing systems},
  2015, pp. 2017--2025.

\bibitem{ssim}
Z.~Wang, A.~C. Bovik, H.~R. Sheikh, and E.~P. Simoncelli, ``Image quality
  assessment: from error visibility to structural similarity,'' \emph{IEEE
  transactions on image processing}, vol.~13, no.~4, pp. 600--612, 2004.

\bibitem{inthewild}
A.~Gordon, H.~Li, R.~Jonschkowski, and A.~Angelova, ``Depth from videos in the
  wild: Unsupervised monocular depth learning from unknown cameras,'' in
  \emph{Proceedings of the IEEE International Conference on Computer Vision},
  2019, pp. 8977--8986.

\bibitem{adam}
D.~P. Kingma and J.~Ba, ``Adam: A method for stochastic optimization,''
  \emph{arXiv preprint arXiv:1412.6980}, 2014.

\bibitem{Kitti1}
A.~Geiger, P.~Lenz, and R.~Urtasun, ``Are we ready for autonomous driving? the
  kitti vision benchmark suite,'' in \emph{2012 IEEE Conference on Computer
  Vision and Pattern Recognition}.\hskip 1em plus 0.5em minus 0.4em\relax IEEE,
  2012, pp. 3354--3361.

\bibitem{Kitti2}
A.~Geiger, P.~Lenz, C.~Stiller, and R.~Urtasun, ``Vision meets robotics: The
  kitti dataset,'' \emph{The International Journal of Robotics Research},
  vol.~32, no.~11, pp. 1231--1237, 2013.

\bibitem{euroc}
M.~Burri, J.~Nikolic, P.~Gohl, T.~Schneider, J.~Rehder, S.~Omari, M.~W.
  Achtelik, and R.~Siegwart, ``The euroc micro aerial vehicle datasets,''
  \emph{The International Journal of Robotics Research}, vol.~35, no.~10, pp.
  1157--1163, 2016.

\bibitem{resnet}
K.~He, X.~Zhang, S.~Ren, and J.~Sun, ``Deep residual learning for image
  recognition,'' in \emph{Proceedings of the IEEE conference on computer vision
  and pattern recognition}, 2016, pp. 770--778.

\bibitem{imagenet}
J.~Deng, W.~Dong, R.~Socher, L.-J. Li, K.~Li, and L.~Fei-Fei, ``Imagenet: A
  large-scale hierarchical image database,'' in \emph{2009 IEEE conference on
  computer vision and pattern recognition}.\hskip 1em plus 0.5em minus
  0.4em\relax Ieee, 2009, pp. 248--255.

\bibitem{pytorch}
A.~Paszke, S.~Gross, F.~Massa, A.~Lerer, J.~Bradbury, G.~Chanan, T.~Killeen,
  Z.~Lin, N.~Gimelshein, L.~Antiga, A.~Desmaison, A.~Kopf, E.~Yang, Z.~DeVito,
  M.~Raison, A.~Tejani, S.~Chilamkurthy, B.~Steiner, L.~Fang, J.~Bai, and
  S.~Chintala, ``Pytorch: An imperative style, high-performance deep learning
  library,'' in \emph{Advances in Neural Information Processing Systems 32},
  H.~Wallach, H.~Larochelle, A.~Beygelzimer, F.~d\textquotesingle
  Alch\'{e}-Buc, E.~Fox, and R.~Garnett, Eds.\hskip 1em plus 0.5em minus
  0.4em\relax Curran Associates, Inc., 2019, pp. 8024--8035.

\bibitem{scsfmlearner}
J.~Bian, Z.~Li, N.~Wang, H.~Zhan, C.~Shen, M.-M. Cheng, and I.~Reid,
  ``Unsupervised scale-consistent depth and ego-motion learning from monocular
  video,'' in \emph{Advances in neural information processing systems}, 2019,
  pp. 35--45.

\bibitem{evo}
M.~Grupp, ``evo: Python package for the evaluation of odometry and slam.''
  \url{https://github.com/MichaelGrupp/evo}, 2017.

\bibitem{ORB1}
R.~Mur-Artal, J.~M.~M. Montiel, and J.~D. Tardos, ``Orb-slam: a versatile and
  accurate monocular slam system,'' \emph{IEEE transactions on robotics},
  vol.~31, no.~5, pp. 1147--1163, 2015.

\end{thebibliography}

\end{document}